\documentclass[conference]{IEEEtran}

\usepackage{graphicx}
\usepackage{color}
\usepackage{balance}

\definecolor{darkgreen}{rgb}{0,0.4,0}

\begin{document}

\title{Machine Learning for Indoor Localization\\Using Mobile Phone-Based Sensors}

\author{\IEEEauthorblockN{David Mascharka and Eric Manley}
\IEEEauthorblockA{Department of Mathematics and Computer Science\\
Drake University\\
Des Moines, Iowa 50311\\
Email: \{david.mascharka, eric.manley\}@drake.edu}}

\maketitle

\begin{abstract}
In this paper we investigate the problem of localizing a mobile device based on readings from its embedded sensors utilizing machine learning methodologies. We consider a real-world environment, collect a large dataset of 3110 datapoints, and examine the performance of a substantial number of machine learning algorithms in localizing a mobile device. We have found algorithms that give a mean error as accurate as 0.76 meters, outperforming other indoor localization systems reported in the literature. We also propose a hybrid instance-based approach that results in a speed increase by a factor of ten with no loss of accuracy in a live deployment over standard instance-based methods, allowing for fast and accurate localization. Further, we determine how smaller datasets collected with less density affect accuracy of localization, important for use in real-world environments. Finally, we demonstrate that these approaches are appropriate for real-world deployment by evaluating their performance in an online, in-motion experiment.
\end{abstract}

\section{Introduction}
As smartphones and other mobile devices become ubiquitous, applications that are able to harness contextual information such as position become increasingly powerful. Uses for indoor localization systems include context-based targeted advertising \cite{context-based-advertising}, emergency response and assisted living \cite{manley}, robotics applications \cite{robot-localization}, and indoor navigation in places such as airports, malls, and campuses.

Machine learning is a field of artificial intelligence dealing with algorithms that improve performance over time with experience. Supervised learning algorithms for regression are trained on data with the correct value given along with each variable. This allows the learner to build a model based on the attributes that best fit the correct value. By giving more data to the algorithm the model is able to improve. Learning can be described in this way as improving performance. The measure of performance is how well the algorithm predicts the regression value given a set of variables or attributes. Machine learning algorithms provide excellent solutions for building models that generalize well given large amounts of data with many attributes by discovering patterns and trends in the data; a task that is often difficult or impossible by other means.

With an increasing number of sensors being made available in the majority of mobile devices, large amounts of data can be collected and used to aid in the localization process. Machine learning algorithms are a natural solution for sifting through these large datasets and determining the important pieces of information for localization, building accurate models to predict an indoor position. Machine learning algorithms may also provide a fast, efficient method for indoor tracking, which will often be more useful to applications than static localization.

In this study, we perform a large-scale analysis of a wide range of machine learning algorithms using real-world data for localization. We also present a hybrid approach suitable for live deployment, which presents a possible solution for algorithms whose calculations are too slow for real-world use. Finally, we conduct an online, in-motion evaluation of the best-performing offline models to show their usefulness when fully deployed in a live, dynamic environment.

The paper is organized as follows. In Section~\ref{sec:related}, we examine other indoor localization systems and their limitations. In Section~\ref{sec:methodology}, we describe our application and data collection process, the testing environment, and how our analysis was conducted. Section~\ref{sec:results} presents the results of our analysis on the full dataset and smaller partial sets of data in an offline environment. This section also presents our hybrid approach to localization and results from our online, in-motion analysis. Finally, the paper concludes with a discussion of future directions for research in Section~\ref{sec:conclusions}.

\section{Related Work}\label{sec:related}
Indoor localization research has garnered a good deal of interest from both academia and industry, with numerous systems being proposed using a variety of technologies. A major disadvantage of many of these systems\textemdash such as infrared \cite{robot-localization}, ultrasound \cite{cricket}, and rfid \cite{rfid, landmarc}\textemdash is that they require dedicated sensors and substantial infrastructure changes and as a result, incur a significant cost to deploy.

Effort has been made to devise localization systems that require little to no infrastructure change using Bluetooth \cite{bluetooth-advertising, bluetooth-localization} and WiFi signal strengths \cite{RADAR, WiGEM, smartphone-localization, HORUS} with some success. The systems developed using WiFi signal strengths for localization show promise but have yet to receive widespread adoption. These systems can be divided into two categories: those using a fingerprinting aproach using algorithms for ``nearest neighbor in signal space'' and those using more complex signal propagation algorithms to determine a device's distance from the access points in range.

Localization systems that use a nearest neighbor in signal space approach require collection of datapoints throughout the room or building they will be deployed in. To predict a position, a new set of attributes constituting a new data point is compared with every point in the classified dataset. Depending on the implementation of the $k$-Nearest Neighbor algorithm, the coordinates of the closest point are used as the coordinates for the new point or an average of $k$ closest points can be used with different weights. These instance-based machine learning approaches can achieve accuracies up to 2 meters on average \cite{smartphone-localization}, but current research is limited in that only one or a few algorithms are considered and do not take into account many of the sensors available in most modern mobile devices. Further, these algorithms are limited by the size of the dataset. A very large dataset will require a substantial amount of time to predict a position, hindering real-world deployment.

Systems that build signal propagation maps for a building have achieved similar accuracy but generally require a great deal of information about the WiFi access points that may not be known or readily available such as their position in the building and their broadcasting power in order to perform localization. Furthermore, the various walls and other objects in a room affect the way in which a signal will propagate. These fluctuations can be difficult to model. A few attempts have been made to remove the requirement of knowledge about access points in range \cite{WiGEM} by using sniffer devices and a centralized localization server. The downside is that localization cannot be performed on the device itself: it must communicate with the server to get a position. Additionally, these models take into account only WiFi signal strengths for localization and so may not be as accurate as models that account for more variables.

In this paper, we expand on the fingerprinting approach described above by exploring a variety of machine learning algorithms using the WiFi access points in range of a mobile device as well as the other sensors embedded in the device for localization and tracking. The model built for an algorithm can be easily implemented as part of an application and installed for localization on the device itself. 

\section{Methodology}\label{sec:methodology}
The localization process consists of two distinct phases: data collection and analysis. Before collecting our dataset, some preprocessing was necessary to determine which WiFi base station IDs to store. In an initial scan of the various signals received throughout the building, we detected a few portable hotspots likely from people in the building tethering their devices. As these signals would not remain constant, they were not included as attributes for the algorithms to train on. Only the WiFi signals that were part of the building infrastructure were stored.

\subsection{Android Application}
We chose to use the Android platform for our research because of the wide variety of devices and sensors available on the market, ease of deployment, and widespread use of the platform.

\begin{figure}[!t]
	\centering
	\includegraphics[keepaspectratio=true,width=0.4\textwidth]{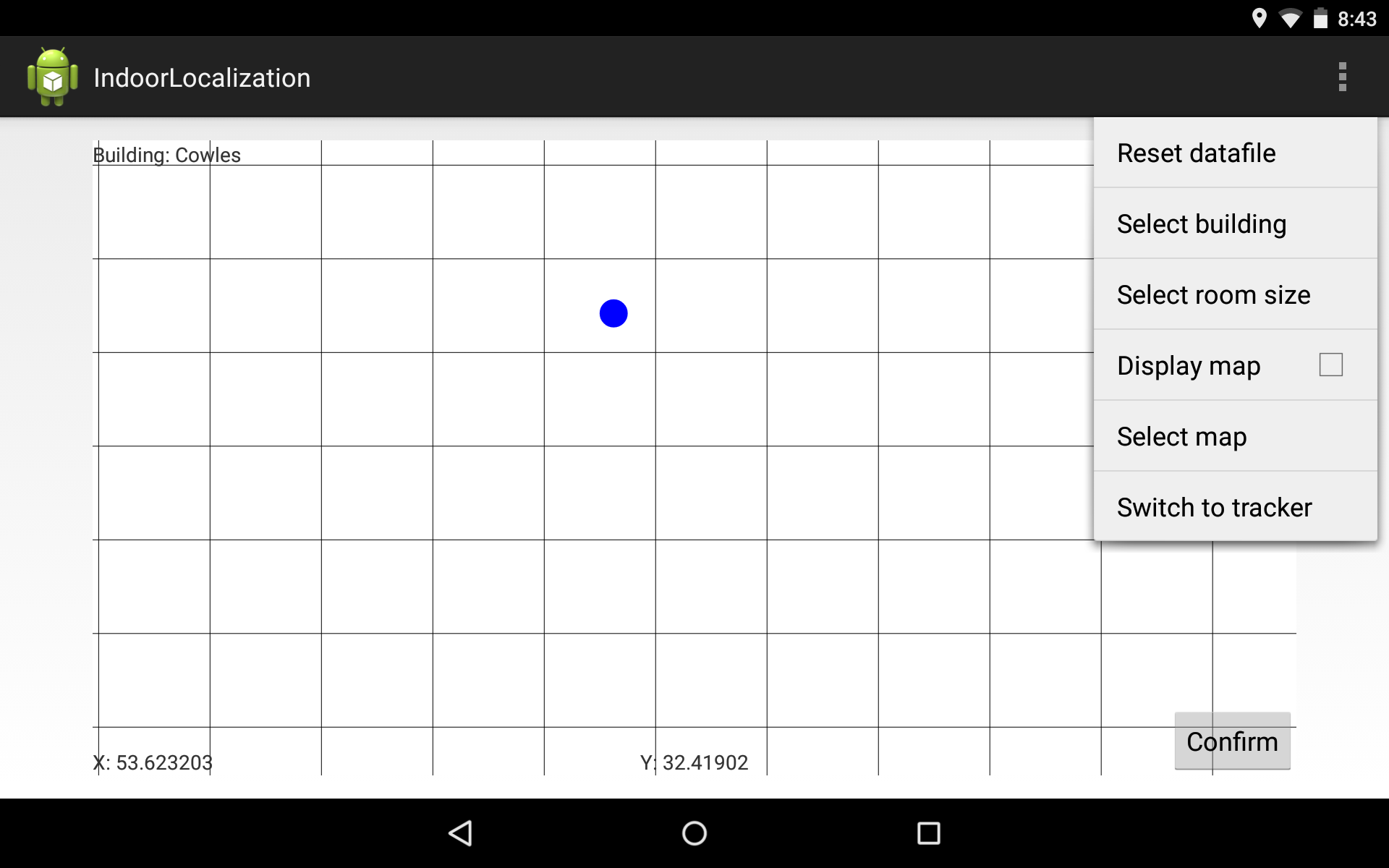}
	\caption{Android application for collecting data}
	\label{fig:app}
\end{figure}

We developed an application that allows a data collector to select which building they are in, allowing for collection of multiple datasets in buildings throughout our campus. Each building has the WiFi access points that will be used programmed into the application and any access points received not in the chosen building's list will be ignored. This allows us to filter out access points that only appear at times such as personal hotspots. The data collector can also select a room or building size and a grid is drawn of the proper size to allow the data collector to more easily indicate their position in the room. Figure~\ref{fig:app} shows the application in the data collection mode.

\subsection{Data Collection}
The data collection phase consists of moving about the building taking readings of the WiFi signal strengths and pulling data from the other sensors in the device. The data is associated with a user-provided location and written to a text file on the device, which can be pulled from the device later for analysis.

A single datapoint, in our case, consists of 172 attributes corresponding to values from each sensor on the device. We took into account a value from the light sensor, GPS/Network location data, signal strengths to 156 WiFi radios from 21 access points, and $x$, $y$, and $z$ values for the device accelerometer, magnetometer, rotation sensor, and orientation sensor. The number of WiFi signals to account for will vary depending on where the localization system will be deployed. While we recorded GPS/Network location data at each point, the accuracy was generally extremely low in our experiments. However, for areas near windows or doors, this may be useful to account for in predicting a position, which motivated us to record it.

In total, we collected 3110 datapoints in the Cowles Library at Drake University in a space about 62 meters wide by 39 meters long (204 feet by 128 feet) using a Motorola XT875. Collection of data involved initiating a scan of WiFi networks in range of the device to record up-to-date signal strengths. When the scan finished, data from all the other sensors in the device were written to a text file with the WiFi signal strengths and the data collector's position in the room, indicated by the data collector in the application itself. This process took an average of five seconds to complete on our device, discounting the data collector's time to indicate their position and move between points in the building.

\begin{figure}[!t]
	\centering
	\includegraphics[keepaspectratio=true,width=0.4\textwidth]{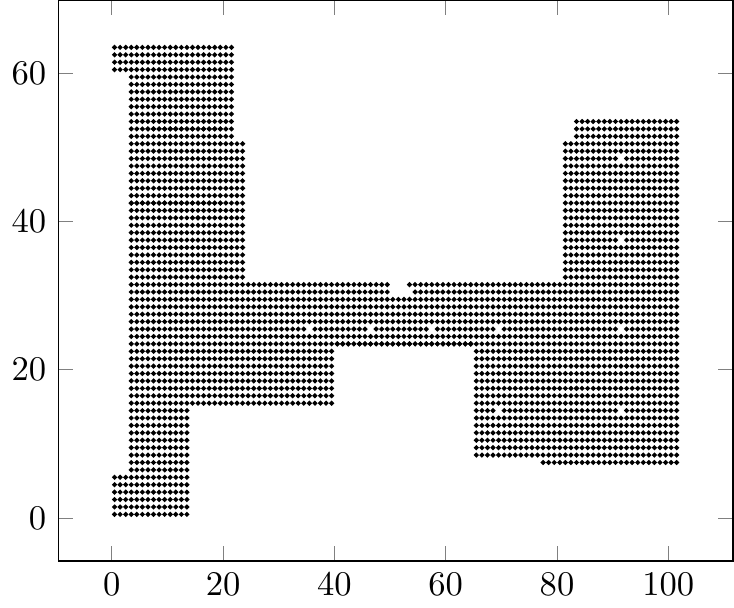}
	\caption{Map of collected datapoints}
	\label{fig:dataplot}
\end{figure}

Measurements were taken in a grid based on the ceiling tiles in the building and all measurements are relative to the building. Each ceiling tile is 2 feet by 2 feet, meaning a tile at coordinates (4, 17) is 8 feet right and 34 feet back from the origin, which in our case was the front left of the building from the entrance. This is an arbitrary measure chosen for ease of use in our case and can be modified to fit any desired building layout. If latitude and longitude coordinates are desired, an implementation can be given the latitude and longitude for diagonal corners of a room and interpolate. The map of collected datapoints can be seen in Figure~\ref{fig:dataplot}. The missing points are caused by obstacles in the building such as pillars, which made it impossible to record data at these points. The left portion of the building contained stacks of books while the rest of the area consisted mostly of open space with tables and desks throughout. We chose to include the area with the stacks of books specifically because it represents a particularly dense set of obstacles in an area that would serve as an excellent challenge to any localization system.

\subsection{Analysis}
After data collection, we transition into a preprocessing and offline analysis phase in which various machine learning algorithms are trained and their errors measured.

Some preprocessing of the data may be necessary or may be helpful depending on the algorithm. In our case, we removed the $x$ attribute when training algorithms to classify $y$ position and removed the $y$ attribute when training for $x$ classification because these will not be known in an online, live environment.

The implementations of the algorithms used was provided by the Waikato Environment for Knowledge Analysis (WEKA) \cite{WEKA}, developed at the University of Waikato. WEKA allows for easy selection of various algorithms and parameter options, which allowed us to test different parameters for each algorithm.

The final phase is an online analysis, in which the algorithms are tested in real-world conditions including with the user in motion, receiving new, unclassified data to process live. Live, in-motion testing is difficult to conduct with accuracy and to our knowledge has not been done to this extent for other similar indoor localization systems. To test our algorithms, we designed a route to walk through the building that would give a representative time series for someone actually traversing the building. While walking, new readings would be taken from the sensors whenever available and given to the algorithms to predict a position. The prediction was output along with a timestamp and written to a text file on the device. A timestamp was also recorded when the researcher reached each vertex of the path and changed directions. The vertex timestamps were used to calculate the researcher's actual position at each time of prediction by interpolating position between vertex timestamps. This position was then compared with the algorithm's prediction to determine the algorithm's error.

\section{Results}\label{sec:results}
Because we were interested in performance not only in a static, offline analysis but in a real-world environment in motion, we split our analysis into two distinct phases: offline and online. The offline analysis section presents the performance of several algorithms in predicting a position. The online analysis section introduces our hybrid approach and shows its performance in live, in-motion testing.
\subsection{Offline}
\subsubsection{Full Dataset}
In total, we examined the performance of 20 well-known machine learning algorithms and 87 different parameter settings on our dataset. All results were obtained using a tenfold cross-validation and verified with ten repetitions. Table~\ref{tab:offline_results} shows the table of results, giving the best mean error for each algorithm. The number in parentheses next to an algorithm name indicates how many variations of the algorithm with different parameters were trained.

\begin{table}[!t]
  \renewcommand{\arraystretch}{1.3}
  \caption{Mean errors in the $x$ and $y$ predictions in the offline analysis}
  \label{tab:offline_results}
  \centering
  \begin{tabular}{|c|c|c|}
    \hline
    Algorithm & Best $x$ Mean Error (m) & Best $y$ Mean Error (m)\\
    \hline\hline
    K* (12) & 1.134 & 0.762\\
    MultiScheme & 1.135 & 0.762\\
    Voting & 1.058 & 1.015\\
    $k$-Nearest Neighbor (22) & 1.417 & 1.227\\
    RBFRegressor (11) & 1.374 & 1.333\\
    RandomForest & 1.454 & 1.470\\
    M5P & 1.671 & 1.429\\
    M5Rules & 1.756 & 1.604\\
    REPTree & 1.790 & 1.665\\
    MLPReg (6) & 1.821 & 2.143\\
    SMOReg (6) & 2.043 & 2.010\\
    RandomTree & 2.150 & 1.950\\
    MultilayerPerceptron (9) & 1.963 & 2.530\\
    DecisionTable & 2.595 & 1.964\\
    RBFNetwork (3) & 2.798 & 3.434\\
    LinearRegression & 3.478 & 3.362\\
    LWL (6) & 5.199 & 4.159\\
    DecisionStump & 6.220 & 4.851\\
    SimpleLinearRegression & 6.655 & 4.822\\
    ZeroR & 20.100 & 7.249\\
    \hline
  \end{tabular}
\end{table}

The algorithm that performed best on our dataset was the K* algorithm \cite{kstar}, an instance-based approach that uses an entropy-based distance function, with mean errors of 1.13 and 0.76 meters for $x$ and $y$ position, respectively, for an absolute mean error of 1.36 meters. Another algorithm that performed well for both $x$ and $y$ classification is the RBFRegressor implementation \cite{rbfreg}, a radial basis function network trained in a fully supervised manner, with mean errors of 1.37 and 1.33 meters, respectively, resulting in an absolute mean error of 1.91 meters. When combining these algorithms using a voting scheme taking the average of both predictions, the $x$ error is reduced to 1.05 meters, while the $y$ error increases slightly from only using K* to 1.01 meters. Using the voting scheme for $x$ classification and K* for $y$, the absolute mean error is reduced from 1.36 to 1.30 meters.

\subsubsection{Partial Datasets}
\begin{figure}[!t]
	\begin{center}
		\includegraphics[keepaspectratio=true, width=0.4\textwidth]{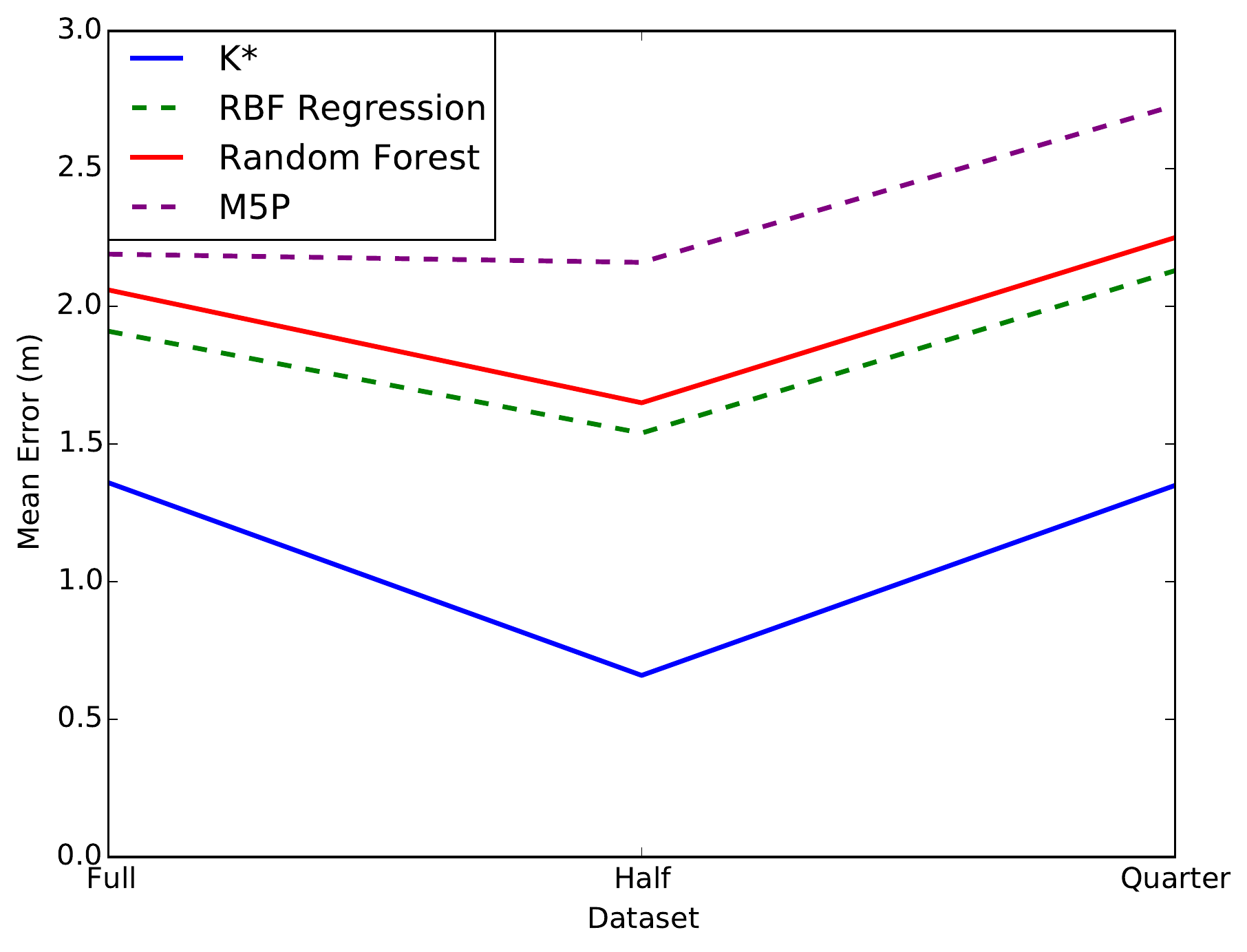}
		\caption{Performance of the most accurate algorithms over each dataset in offline analysis}
		\label{fig:comparison}
	\end{center}
\end{figure}

We looked not only at how the algorithms performed on the full dataset but were also interested in how performance may deteriorate as less data is used for training. This information will be useful moving into real-world deployment as less data needed to implement a localization system means faster and easier deployment. We examined algorithm accuracy using one-half the full dataset and one-quarter the full dataset, with both sets of data in a grid pattern like the full dataset. Table~\ref{tab:half} shows the results using one-half the data, while the results for one-quarter can be seen in Table~\ref{tab:quarter}. A comparison of the four most accurate algorithms over each dataset can be seen in Figure~\ref{fig:comparison}. The full dataset had a density of one reading every two square feet. The one-half dataset was built by taking every second reading from the full dataset while the one-quarter dataset was built by taking every fourth reading from the full set.

The error using half the original dataset results in slightly increased accuracy for most of the tested algorithms. The K* error is reduced to 0.563 meters in $x$ prediction and 0.395 meters in $y$ prediction, resulting in an absolute error of 0.763 meters. The RBF algorithm's error is also slightly reduced to 1.113 meters and 1.053 meters for $x$ and $y$ prediction respectively, which gives an absolute error of 1.532 meters. This slight reduction leads us to believe that the full dataset may have a density that results in overfitting and too little variation between sets of readings, leading to difficulty predicting a correct position. It may also be the case that the RBF model performs better using half the data because the network built on the full set was too small to completely capture the complexity of the data.

The errors when using only one quarter of the original dataset increased slightly on average, up to almost half a meter. The K* error decreased slightly in the $x$ prediction and increased slightly in the $y$ prediction to 1.077 meters and 0.831 meters respectively, resulting in an absolute error of 1.36 meters, nearly unchanged from the full dataset. The RBF regression model increased in both $x$ and $y$ predictions almost one fifth of a meter to 1.561 meters and 1.511 meters respectively, increasing the absolute error to 2.173 meters, about 0.3 meters higher than the full dataset. These results indicate that this density of readings may not be high enough for applications that require a very high degree of accuracy. However, if only a general position is required, a small dataset may allow for faster and easier deployment and may be more reasonable for widespread adoption. A further decline in accuracy is expected if a lower density dataset is used.

\begin{table}[!t]
	\renewcommand{\arraystretch}{1.3}
	\caption{Mean errors in $x$ and $y$ position for half the data with differences from the full set. Better performance compared with the full dataset is indicated by green with worse performance in red.}
	\label{tab:half}
	\centering
	\resizebox{\linewidth}{!}{
	\begin{tabular}{|c|c|c|c|}
		\hline
		Algorithm & Mean $x$ Error (m) & Mean $y$ Error (m) & $x$/$y$ Difference (m)\\
		\hline\hline
		K* & 0.563 & 0.395 & \color{darkgreen}{-0.572}\color{black}/\color{darkgreen}{-0.368}\\
		$k$-Nearest Neighbor & 0.695 & 0.841 & \color{darkgreen}{-0.723}\color{black}/\color{darkgreen}-0.387\\
		RBFRegressor & 1.113 & 1.053 & \color{darkgreen}-0.26\color{black}/\color{darkgreen}-0.280\\
	    RandomForest & 1.146 & 1.194 & \color{darkgreen}{-0.308}\color{black}/\color{darkgreen}-0.277\\
	    M5P & 1.450 & 1.603 & \color{darkgreen}{-0.221}\color{black}/\color{red}+0.173\\
	    M5Rules & 1.597 & 1.847 & \color{darkgreen}{-0.159}\color{black}/\color{red}+0.242\\
	    REPTree & 1.707 & 1.768 & \color{darkgreen}{-0.084}\color{black}/\color{red}+0.102\\
	    SMOReg & 1.250 & 1.268 & \color{darkgreen}{-0.794}\color{black}/\color{darkgreen}-0.742\\
	    RandomTree & 1.292 & 1.195 & \color{darkgreen}{-0.858}\color{black}/\color{darkgreen}-0.755\\
	    MultilayerPerceptron & 1.743 & 2.176 & \color{darkgreen}{-0.221}\color{black}/\color{darkgreen}-0.355\\
	    DecisionTable & 2.707 & 2.560 & \color{red}{+0.112}\color{black}/\color{red}+0.596\\
	    RBFNetwork & 2.859 & 2.877 & \color{red}+0.615\color{black}/\color{darkgreen}-0.558\\
	    LinearRegression & 3.414 & 3.426 & \color{darkgreen}-0.064\color{black}/\color{red}+0.063\\
	    LWL & 5.157 & 5.456 & \color{red}+0.317\color{black}/\color{red}1.297\\
	    DecisionStump & 6.242 & 6.267 & \color{red}+0.021\color{black}/\color{red}+1.415\\
	    SimpleLinearRegression & 6.651 & 6.651 & \color{red}+0.094\color{black}/\color{red}+1.829\\
	    ZeroR & 20.092 & 20.086 & \color{darkgreen}-0.0085\color{black}/\color{red}+12.837\\
		\hline
	\end{tabular}
	}
\end{table}

\begin{table}[!t]
	\renewcommand{\arraystretch}{1.3}
	\caption{Mean errors in $x$ and $y$ position for one-quarter the data with differences from the full set. Better performance compared with the full dataset is indicated by green with worse performance in red.}
	\label{tab:quarter}
	\centering

	\resizebox{\linewidth}{!}{
	\begin{tabular}{|c|c|c|c|}
		\hline
		Algorithm & Mean $x$ Error (m) & Mean $y$ Error (m) & $x$/$y$ Difference (m)\\
		\hline\hline
		K* & 1.077 & 0.831 & \color{darkgreen}-0.058\color{black}/\color{red}+0.068\\
		$k$-Nearest Neighbor & 1.396 & 1.475 & \color{darkgreen}-0.022\color{black}/\color{red}+0.247\\
		RBFRegressor & 1.561 & 1.511 & \color{red}+0.187\color{black}/\color{red}+0.178\\
		RandomForest & 1.567 & 1.622 & \color{red}+0.113\color{black}/\color{red}+0.151\\
		M5P & 1.823 & 2.030 & \color{red}+0.152\color{black}/\color{red}+0.600\\
		M5Rules & 2.067 & 2.118 & \color{red}+0.311\color{black}/\color{red}+0.513\\
		REPTree & 2.249 & 2.048 & \color{red}+0.458\color{black}/\color{red}+0.382\\
		SMOREG & 2.073 & 2.067 & \color{red}+0.029\color{black}/\color{red}+0.057\\
		RandomTree & 2.085 & 2.115 & \color{darkgreen}-0.065\color{black}/\color{red}+0.165\\
		MultilayerPerceptron & 2.524 & 3.097 & \color{red}+0.560\color{black}/\color{red}+0.566\\
		DecisionTable & 2.743 & 2.975 & \color{red}+0.148\color{black}/\color{red}+1.011\\
		RBFNetwork & 2.932 & 3.158 & \color{red}+0.133\color{black}/\color{darkgreen}-0.277\\
		LinearRegression & 3.456 & 3.463 & \color{darkgreen}-0.022\color{black}/\color{red}+0.100\\
		LWL & 5.395 & 5.139 & \color{red}+0.195\color{black}/\color{red}+0.980\\
		DecisionStump & 6.248 & 6.279 & \color{red}+0.027\color{black}/\color{red}+1.427\\
		SimpleLinearRegression & 6.626 & 6.632 & \color{darkgreen}-0.030\color{black}/\color{red}+1.810\\
		ZeroR & 20.086 & 20.09 & \color{darkgreen}-0.015\color{black}/\color{red}+12.841\\
		\hline
	\end{tabular}
	}
\end{table}

\subsection{Online}
In the online phase we saved the models for the two best-performing algorithms on the full dataset to test, examining the K* and RBF regression algorithms. In total we collected 27 sets of test results for K* and 20 for the RBF model. The bulk of these results were collected several months after our initial data collection, indicating some stability of the original data and algorithms.

In a live setting, the difference in how these two algorithms work is very important. As K* is instance-based, it compares each new datapoint to every classified point in the dataset. In contrast, the RBF algorithm learns weights for each of the 172 attributes in the data and must only multiply these weights by the attribute values of a new datapoint, then add these together to predict a position. This is a much faster operation than the entropy calculation for each of the 3110 points in the dataset and makes a substantial difference in a live environment. Initially, K* took 30 to 45 seconds to calculate a position; much too long for real-world applications. In contrast, the RBF model predicts a location almost instantly.

\subsubsection{Proposed Hybrid Approach}
\begin{figure}[!t]
	\centering
	\includegraphics[keepaspectratio=true, width=0.4\textwidth]{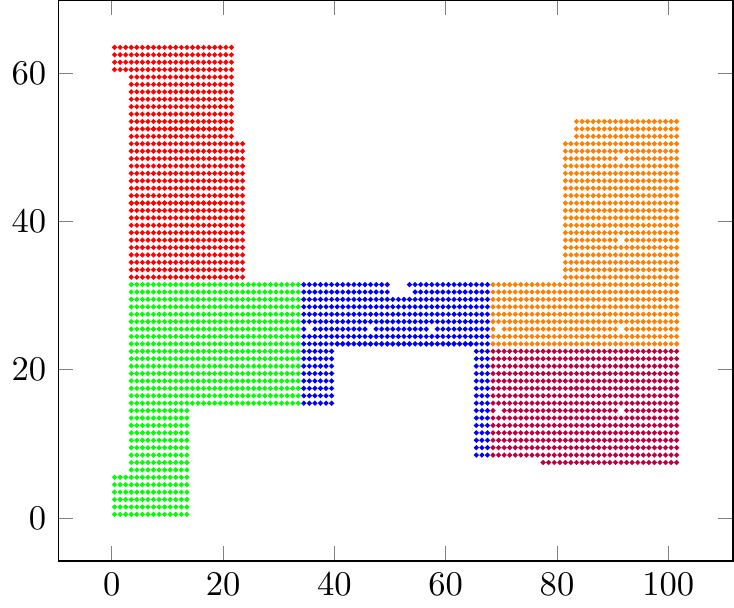}
	\caption{Partitioned dataset, where each color corresponds to a different partition}
	\label{fig:partition}
\end{figure}

To solve the time problem for K* we decided to break our full dataset into smaller partitions and trained K* classifiers on each partition. Our partitioning can be seen in Figure~\ref{fig:partition}. To determine which partition of the building the user was in, and thus which K* classifier to use to calculate the user's position, we again looked at machine learning methods, settling on a random forest model \cite{randomforest} which achieved over 96\% accuracy in a tenfold cross-validation. This reduced the number of comparisons from 3110 to about 400 to 500 and substantially increased the speed from 30-45 seconds to 3 seconds: fast enough to be useful in a real-world setting. A hybrid approach such as this may also speed up other instance-based localization systems, which will take too long to be useful with large datasets otherwise.

\subsubsection{Online, Static Results}
We were interested first in how the algorithms would perform with a user standing still at a point in the building. K* achieved accuracies within three meters at every point, with most predictions within one meter of the user's actual position. The RBF regression algorithm performed similarly in a static online test.

\subsubsection{Online, In-motion Results}
\begin{figure}[!t]
  \begin{center}
    \includegraphics[keepaspectratio=true, width=0.4\textwidth]{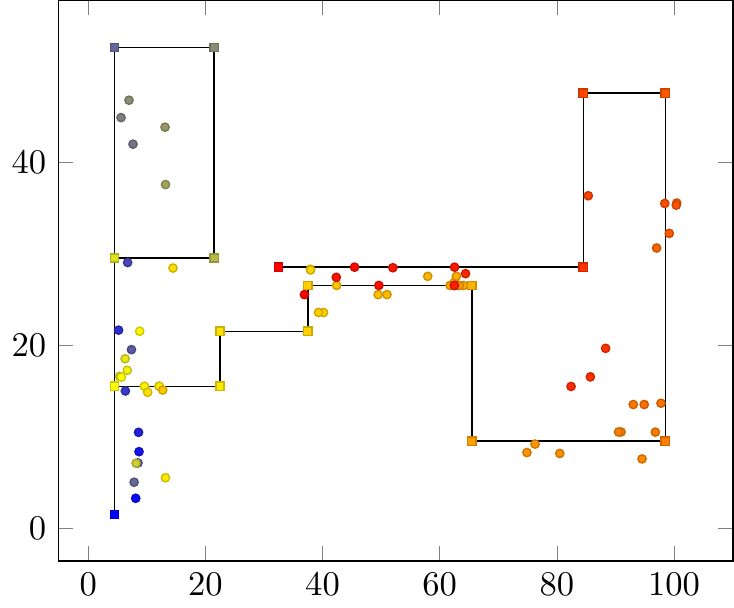}
 	\caption{Live Test: K* algorithm predicting position as the researcher walks the path in black allowing the device to change orientation. Color indicates time starting at blue, ending at red with smooth interpolation between.}
    \label{fig:live}
  \end{center}
\end{figure}

After the static testing, we were interested in the performance of both algorithms in an online, in-motion testbed, mimicking a real-world environment. We collected datasets with the user walking at various paces along the planned route to determine whether speed of movement affected accuracy. We looked at a normal pace of about 1.15 meters per second, a slow pace of 0.75 meters per second, and a quick pace of 1.69 meters per second. We also wanted to determine whether changing the device orientation would affect accuracy. Since data was collected with the device held facing south, we walked some routes constantly holding the device facing south. We also walked routes allowing the device to change orientation as we walked, mimicking a real-world user. These tests were conducted at each pace. The table of results can be seen in Table~\ref{tab:online}. A single walked route using K* at a slow pace with the device held with changing orientation can be seen in Figure~\ref{fig:live}. The starting point is at $x$ coordinate 4.5, $y$ coordinate 1.5 in the bottom left of the plot and end point $x$ coordinate 32.5, $y$ coordinate 28.5 near the middle in red. The black line indicates the path walked with square points as vertices. The circle points in the plot indicate predicted positions. Color is used to indicate time, interpolated smoothly from blue to red.

\begin{table}[!t]
  \renewcommand{\arraystretch}{1.3}
  \caption{Online, in-motion results showing each algorithm's average error at each pace with different orientations}
  \label{tab:online}
  \centering
  \begin{tabular}{|c|c|c|c|}
    \hline
    Algorithm & Pace & Orientation & Average Error (m)\\
    \hline\hline
    K* & Slow & Constant & 6.03\\
    K* & Slow & Changing & 5.08\\
    K* & Normal & Constant & 6.39\\
    K* & Normal & Changing & 6.23\\
    K* & Fast & Constant & 9.19\\
    K* & Fast & Changing & 8.53\\
    RBF & Slow & Constant & 7.85\\
    RBF & Slow & Changing & 7.39\\
    RBF & Normal & Constant & 8.86\\
    RBF & Normal & Changing & 9.21\\
    RBF & Fast & Constant & 10.15\\
    RBF & Fast & Changing & 9.85\\
    \hline
  \end{tabular}
\end{table}

Allowing the device to change orientation as the user walks does not appear to significantly affect accuracy, even though data was collected in only one device orientation. This indicates that orientation of the device likely plays an insignificant role in localization compared with other attributes. Removing this attribute may help increase accuracy as algorithms would have fewer attributes to build a model on.

It is significant that the testing was done at the library at Drake University. The left area of the live testing was amidst densely-packed bookshelves. Given that wireless signal strengths contributed most to the calculated position and the bookshelves likely affected signal propagation a great deal this area tended to have the highest error. Removing the positions in this area from our calculations improved accuracy an average of one meter, which may be more indicative of a normal indoor environment.

\section{Conclusions and Future Work}\label{sec:conclusions}
In this work, we have examined a large number of machine learning algorithms for indoor localization based on the sensors readily available in smartphones. We have found algorithms giving an accuracy up to 0.76 meters on average in a real-world environment without the need for dedicated hardware or changes to infrastructure, outperforming algorithms considered in previous studies \cite{smartphone-localization}.  Our online, in-motion experiments show that our proposed hybrid approach achieved accuracy on par with the best offline instance-based methods with the speed of non-instance-based methods.

In the future, we will explore using multiple devices to collect and evaluate models on. As different devices may receive different signal strengths from each access point at various locations and may receive a different number of signals depending on whether the device possesses a single- or dual-band wireless card, a model built from a dataset from one device may not accurately predict another device's location. Some small-scale experiments have shown promise in collecting datasets from multiple devices and aggregating them, then allowing the algorithms to build a model from this combined dataset.

We also intend to look into how to determine the required density for achieving a desired level of accuracy and determining an optimal pattern for collecting data. This may reduce the number of necessary datapoints and improve accuracy. Other areas to investigate include taking multiple readings at different times, which may improve accuracy and reliability since signal strengths may fluctuate throughout the day. 
\balance
Futhermore, the increased error moving from an offline to online environment may be due to the time taken to calculate a position. In the case of K*, this may be due to the time taken to calculate a position. Because K* takes an average of three seconds to calculate a position and the user is moving during the calculation, the predicted position may be indicative of the user's position seconds in the past by the time computation finishes. Research is ongoing to determine whether this is the case. If so, we will focus on determining the best method to reduce this error, looking at incorporating user velocity and direction using modified methods for dead reckoning to achieve the best possible accuracy.

\bibliographystyle{IEEEtran}
\bibliography{references}

\end{document}